\documentclass[conference]{IEEEtran}
\IEEEoverridecommandlockouts
\usepackage{cite}
\usepackage{amsmath,amssymb,amsfonts}
\usepackage{graphicx}
\usepackage{textcomp}
\usepackage{xcolor}
\usepackage{graphicx}
\usepackage[utf8]{inputenc} 
\usepackage[T1]{fontenc}    
\usepackage{hyperref}      
\usepackage{url}           
\usepackage{booktabs}      
\usepackage{amsfonts}      
\usepackage[normalem]{ulem}
\usepackage{nicefrac}      
\usepackage{microtype}      
\usepackage{balance}
\usepackage{mathtools}
\usepackage{scalerel}
\usepackage[bottom]{footmisc}
\usepackage{url}
\usepackage{epsfig}
\usepackage{enumitem}
\usepackage{bm}
\usepackage{arydshln}
\usepackage{amsmath}
\usepackage{epsfig}
\usepackage{pdfsync}
\usepackage{amsmath,amssymb, amstext}
\usepackage{float}
\usepackage{xcolor}
\usepackage{amsmath}
\usepackage{algorithm}
\usepackage{algorithmicx}
\usepackage{algpseudocode}
\def\BibTeX{{\rm B\kern-.05em{\sc i\kern-.025em b}\kern-.08em
    T\kern-.1667em\lower.7ex\hbox{E}\kern-.125emX}}
\usepackage{tikz}
\usepackage{lipsum}

\newcommand\copyrighttext{%
  \footnotesize This paper has been accepted by the IEEE International Conference on Communications (ICC), 28 May–1 June 2023, Rome, Italy.\\ \copyright 2023 IEEE. Personal use of this material is permitted. Permission from IEEE must be obtained for all other uses, in any current or future media, including reprinting/republishing this material for advertising or promotional purposes, creating new collective works, for resale or redistribution to servers or lists, or reuse of any copyrighted component of this work in other works.}
\newcommand\copyrightnotice{%
\begin{tikzpicture}[remember picture,overlay]
\node[anchor=south,yshift=10pt] at (current page.south) {\fbox{\parbox{\dimexpr\textwidth-\fboxsep-\fboxrule\relax}{\copyrighttext}}};
\end{tikzpicture}%
}
\begin{document}

\title{Satellite Anomaly Detection Using Variance Based Genetic Ensemble of Neural Networks\\
\thanks{\textsuperscript{$\ddagger$} Corresponding author.~This work was supported by the High-Throughput and Secure Networks Challenge program of National Research Council Canada under Grant No. CH-HTSN-418.
}
}

\author{\IEEEauthorblockN{
Mohammad Amin Maleki Sadr\IEEEauthorrefmark{1},
Yeying Zhu\IEEEauthorrefmark{1}, and 
Peng Hu\IEEEauthorrefmark{2}\IEEEauthorrefmark{1}{\textsuperscript{$\ddagger$}}}
\IEEEauthorblockA{\IEEEauthorrefmark{1}University of Waterloo, ON N2L 3G1, Canada}
\IEEEauthorblockA{\IEEEauthorrefmark{2}
National Research Council of Canada, Waterloo, ON N2L 3G1, Canada}
\{mohammadamin.malekisadr, yeying.zhu\}@uwaterloo.ca, 
peng.hu@nrc-cnrc.gc.ca
}

\maketitle
\copyrightnotice

\begin{abstract}
In this paper, we use a variance-based genetic ensemble (VGE) of Neural Networks (NNs) to detect anomalies in the satellite's
historical data. We use an efficient ensemble of the predictions from  multiple Recurrent Neural Networks (RNNs) by leveraging each model's
uncertainty level (variance).   For prediction, each RNN is guided by a Genetic Algorithm (GA) which constructs the optimal structure for each RNN model.  However, finding the model uncertainty level is
challenging in many cases. Although the Bayesian NNs (BNNs)-based
methods are popular for providing the confidence bound of the
models, they cannot be employed in complex NN structures as they
are computationally intractable. This paper uses the Monte Carlo (MC) dropout as an approximation version of BNNs. Then these uncertainty
levels and each predictive model suggested by GA are used to generate a new
model, which is then used for forecasting the TS and AD.
Simulation results show that the forecasting and AD capability of
the ensemble model outperforms existing approaches.
\end{abstract}


\section{Introduction}
Satellites suffer from abnormal and unexpected issues. The root causes of these issues can be various radiations from the space atmosphere, human errors, etc. These anomalies  affect the overall system performance degradation \cite{li2019intelligent}.  
The time-series (TS) anomaly detection (AD) approaches proposed in  \cite{hundman2018detecting,  malhotra2015long,su2019robust,geiger2020tadgan,li2019mad} hardly suggest a general framework for neural network (NN)  structure. The genetic algorithm (GA) provides  hyperparameter tuning and an optimal structure for an NN and imposes an extra layer of computations over a NN model.
In this paper, we leverage  GA for fine-tuning the various Recurrent Neural Network (RNN) models. After that, we propose an innovative ensemble approach  to merge the predictions of various  models based on the associated variances.  The main motivation of this paper is to enhance the prediction accuracy as well as precise AD by using  ensembles of various RNNs. 

 In this paper, to build an efficient ensemble model, we utilize the confidence interval for each of the fitted models.       
In the context of Deep Neural Network (DNN), the probability that the  NN predictions drop within a given region around the accurate value is considered as an uncertainty region, and the principal technique to produce this probability is the Bayesian Neural Network  (BNN)  \cite{loquercio_segu_2020}. However, calculating the posterior distribution  of BNN is challenging. For solving this problem, many studies proposed a tractable approximation  for calculating the posterior distribution \cite{ Gal2015,9626568}. 
Among these approaches, Monte Carlo (MC) dropout method in \cite{Gal2015} shows  several advantages over the other approaches including lower complexity and  higher accuracy. In \cite{Gal2015} and \cite{gal2016theoretically}, the authors  employed the MC dropout for approximating BNN at the dense NN and RNN, respectively. 

To capture the confidence region of the normal points, in our proposed NN model consisting of different Long Short-Term Memory (LSTM),  Gated Recurrent Unit (GRU), RNN, and dense layers, we use an extended version of the  MC dropout method in \cite{gal2016theoretically} as an approximated version of Bayesian RNN. By leveraging the statistics of the NN model  resulted from the MC dropout method, we can detect the anomalies. 
Moreover, our variance-based genetic ensemble (VGE) approach has five stages: i) we perform a novel prepossessing technique to the raw data to attain more smoothed data to feed the RNNs;  
ii)  we employ  GA for tuning  different combinations of RNN models, i.e.,  LSTM, simple RNN, and GRU and then select the best model among them; iii) We extract the prediction and  confidence region  of the best-predicted model. Finding the variance of the model for RNN-based approaches is challenging. In this work, we employ the MC dropout method as a tight estimation version of BNN to capture the uncertainty level of the RNN network \cite{gal2016theoretically,Gal2015};  
iv) we combine the mean and variance of the best RNN models to find a more accurate model; 
and v) using the results of the  ensemble model, we perform  post-processing to find the anomaly points. 

Two typical telemetry datasets from NASA \cite{hundman2018detecting} for two different spacecraft are used in our study: Curiosity Rover on Mars (MSL) and Soil Moisture Active Passive satellite (SMAP). The datasets can be considered representative to be applied in generic fault detection, identification, and recovery (FDIR) missions for satellites using the similar features. Table \ref{my_label12e} describe the dataset summary. For each specific time step of the measured data, we have a vector of features including several encrypted commands \cite{hundman2018detecting}. 

\begin{figure}
    \centering
    \includegraphics[width=\linewidth]{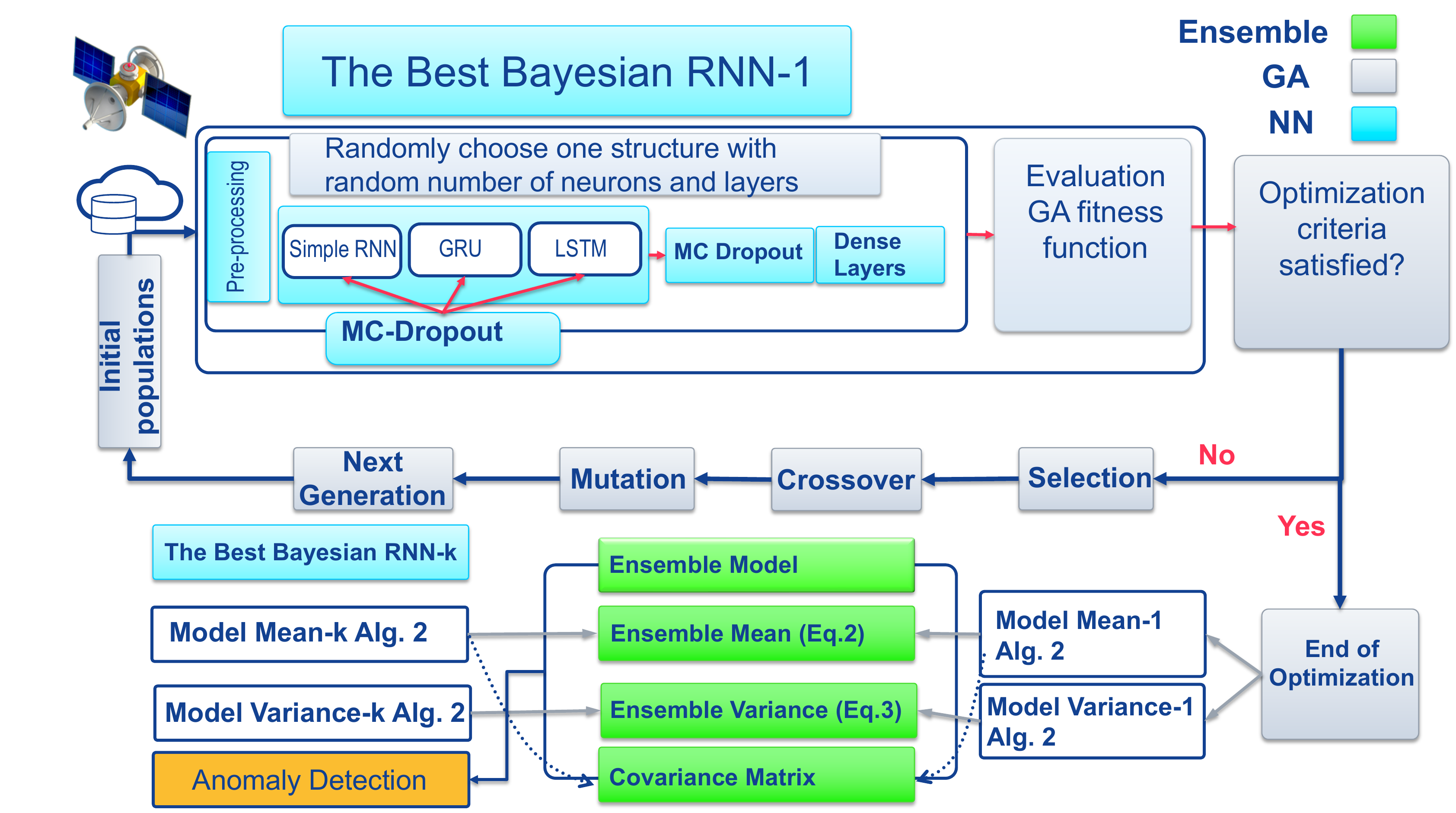}
    \caption{VGE approach overview}
    \label{fig:my_label2}\vspace{-28pt}
\end{figure}

 The main novelties of this paper are summarized as follows:
\begin{itemize}
    \item We propose a novel ensemble approach for combining different RNN models. More specifically, we use the combination of different RNN variations based on their associated uncertainty levels. The models with larger uncertainty regions will get smaller weights during the combinations. The uncertainty region is found for a NN consisting of different combinations of simple RNN, LSTM, GRU, and dense layers. The mathematical formulations of \cite{gal2016theoretically} to cover a general form of Bayesian NN is derived.  
    \item GA is used to find the best model structure,  where we have searched for the best NN structure for each base model with tuned hyperparameters. In Fig. \ref{fig:my_label2}, we show how to use GA for Bayesian RNN.
    \item  One of the main issues in training the NN is over-fitting. As our BNN approach leverages probabilistic modeling and adds a regularization term to the loss function, it does not have this issue \cite{Gal2015}.
\item Our pre-processing approach which uses smoothing the data is also novel. Different from our method in \cite{2022}, in which we used binary coefficients to merge the data we use the continuous  and more efficient method here. 
\end{itemize}

The rest of this paper is structured as follows: Section II introduces the
related works; 
Section III details the pre-processing method; Section IV details the problem and our proposed VGE approach; 
and Section V presents the experimental results,  followed by conclusions in Section VI. 

\renewcommand{\baselinestretch}{0.9}{
\begin{table}
    \caption{ Datasets summary}
   \scalebox{0.75}{ 
   \begin{tabular}{|c|c|c|c|c|}
    \hline
   Dataset&Signals &   Num. of Dimensions & Num. of Data Points & Num. of Anomaly Points\\ \hline
      SMAP  &55 & 25 & 562800&54696 \\ \hline
      MSL &27& 55&  132046 & 7766\\ \hline
    \end{tabular} 
    }  
    \label{my_label12e}
\end{table}
}

\section{Related Works}
The AD literature is extensive and it is beyond the scope of this paper \cite{chandola2009anomaly}.  AD in this context is considered to be unsupervised learning
\cite{chandola2009anomaly,hundman2018detecting}. 
The AD methods can be  categorized into two folds, which are reconstruction and forecasting-based approaches.  
In forecasting-based approaches,  the model predicts the next step $t+1$, based on the sets of  observations ${\cal{T}}={0,1,\dots,t}$.  
If the gap between the predicted and current values becomes  higher than a threshold level, the point is considered a tentative anomaly. Statistical approaches such as ARIMA \cite{pena2013anomaly},   and distance-based approach in \cite{breunig2000lof} are all forecasting-based approaches.  
However, the aforementioned methods may not be robust against the noise level and thus may not have a successful AD  
when the noise is higher than a specific threshold.  On the other hand, another subcategory of forecasting-based methods which are robust against noise level is 
  RNN-based methods.    
  The   RNN-based approaches such as LSTM, GRU, and simple RNN \cite{malhotra2015long} are shown to outperform the traditional statistical-based approaches for AD.   
In \cite{nanduri2016anomaly}, the implementation of both LSTMs and GRUs for AD is discussed. The later work has been extended to satellite applications in  \cite{hundman2018detecting}. 
In \cite{hundman2018detecting}, the authors use  a dynamic thresholding approach based on the GRU and LSTMs for AD in satellite and rover data.

Different from forecasting-based AD, the  reconstruction-based TS AD approaches are performed in offline modes. In this approach, the TS will be rebuilt first, and then based on the reconstruction error the anomalies are flagged. There are several reconstruction-based approaches such as TadGAN \cite{geiger2020tadgan}, MadGan \cite{li2019mad}, and OmniAnomaly \cite{su2019robust}.
 Despite having a good performance in AD,  
 non of the previous studies present a general framework for both  hyperparameters selection  of NN  and AD. GA is used for learning the best hyperparameters and  the optimal architecture \cite{pietron2021fast}.     
   In \cite{faber2021ensemble}, the authors  proposed a fully automated  framework that uses GA for AD purposes. Recently, in  \cite{lyu2022one}, the authors use GA for  automatically designing and training new RNNs for forecasting TS. 

Furthermore,  for improving NN performance,  ensemble learning has been proposed in different studies  \cite{9825269}. 
The key idea of ensemble learning is that the insight from a group of machine learning algorithms is more than a single algorithm.  
In \cite{9825269}, 
a comprehensive survey on ensemble methods is presented.  
While previous works addressed the ensemble of predictions of several models by model-averaging, they have not considered the correlation between different models or quantified each model's uncertainty. For these reasons, the existing ensemble methods can be improved by considering each model's uncertainty level. 

\section{ Pre-processing}

Assume a general multivariate TS data ${\bf{Z}} \in \mathbb{R}^{N \times T}$, where $N$ and $T$ denote the number of channels and the length of the 
TS, respectively. 
we use a sliding window with length ${m}$ to split the original multivariate TS into multiple slices with a pre-defined window length ${m}$, i.e.,  ${\bf{Z}}(m)$,.., ${\bf{Z}}(T)$, and ${{\bf{Z}}(t)} = [{{\bf{z}}(i)}]_{i = t - m + 1}^t$ where ${{\bf{z}}(i)} \in {\mathbb{R}^{N \times m}}$.  
In our pre-processing step, in
order to make the signal  smoother, we implement an innovative windowing-based averaging approach.  Making the data smoother helps better prediction and avoids  a rush decision for flagging the anomalies.   The method works based on adaptive variable length windowing. The input and output of our algorithm are  ${\bf Z} $ and ${\bf{x}}(t) \in  \mathbb{R}^{N \times T}$, respectively.   
By selecting $m$ samples from the input signal, and considering  $\gamma _j(t)$ ($0 \le {\gamma _j}(t) \le 1$) as the weights of the $j$th point inside the $t$th windows, we have 

\begin{equation}
\label{eqq7}
{\bf{x}}(t) = \alpha(t)\sum\limits_{j =  - \frac{{{m(t)}}}{2}}^{\frac{{{m(t)}}}{2}} {{\gamma _j}(t){\bf{z}}(t - j)} 
\end{equation}

where $\alpha(t)=\frac{1}{{\left\| {\boldsymbol{\lambda}}(t) \right\|}}$, $\boldsymbol{\lambda} (t) = {[ {\begin{array}{*{20}{c}}
{{\gamma _{  \frac{{-{m(t)}}}{2}}}}& \ldots &{{\gamma _{\frac{{{m(t)}}}{2}}}}
\end{array}}]^T}$ indicates a vector of weights inside  the $t$th windows. We denote the vector norm by ${\left\| . \right\|}$. We then  define  ${\bf{e}}(t-j)=\left| {{{{\bf{z}}(t-j-1) - {\bf{z}}(t -j )}}} \right|$ as a distance measure between two consecutive points. By initializing  
$m(0)=2$ and ${\gamma _0(0)=1}$, if  we do not notice any abrupt behavior, i.e., ${\bf{e}}(t-j)<e_{th}^j$ where $e_{th}^j$ is a  pre-selected threshold defined as ${e_{th}^j} = \mu_j+ 2\sigma_j$ where $\mu_j$ and $\sigma_j$ are 
the mean and standard deviation of the observations inside the $t$th window with length $m(t)$, respectively.   We increase the window length ($m(t)$=$m(t)+1$) and put  ${\gamma _{ j}}=e^{-{\bf{e}}(t-j)}$. If not, i.e.,   ${\bf{e}}(t-j)\ge e_{th}^j$, we move to the next window, i.e., $t=t+1$.  We do this process for all $ t \in \{ 1,...,T\}$. Here,  
${\bf x}(t)$ acts as the input of the
forecasting model.  
Note that, we present a similar method in \cite{2022}, with a difference that  ${\gamma _{j}}$  is a continuous variable here whereas  in \cite{2022} it was binary. 
\renewcommand{\baselinestretch}{0.85}{
\begin{algorithm}
\hspace*{\algorithmicindent} Assume that for the $t$th window, ${\bf{z}}(t)$ is the input vector and  ${\bf{x}}(t)$ is the output vector of the pre-processing algorithm. Also at the $t$th windows, we have ${\cal{I}}=[ - \frac{{{\bf{m}}(t)}}{2},...,\frac{{{\bf{m}}(t)}}{2}]$ points.  We take the following steps: 
\hspace*{\algorithmicindent} 
\begin{algorithmic}[1]\small
\caption{Pre-processing algorithm}
\State{$m(0)=2$, ${\gamma _0=1}$, $t=0$}
\While { $t \le T$ }
\State{Compute $\mu^{(t)}$, $\sigma^{(t)}$ and then $e_{th}^{(t)} = \mu^{(t)} + 2\sigma^{(t)}$}
\For{$j \in {\cal{I}}$} 
\If{ ${{\bf{e}}(t-j)}<e_{th}^{(t)}$}
\State{${\gamma _{j}}=e^{-{\bf{e}}(t-j)}$}
\State{$m(t)=m(t)+1$}
\ElsIf{ ${{\bf{e}}(t-j)}>e_{th}^{(t)}$}
\State{\textit{Break}}
\EndIf
\EndFor
\State{ Compute (\ref{eqq7})}
\State{ $t=t+1$}
\EndWhile
\end{algorithmic}
\end{algorithm}}

\section{Problem Statement}
\subsection{Ensemble Models} 
Assume that we have $K$ RNN models and the predicted value by the $j$th model is ${ \bf{\hat y}} _j$. Let  ${w_j} = \frac{1}{{\sigma _j}}$ for $j \in {\cal{P}}$ where ${\cal{P}}=\{1,...,K\}$ is the inverse variance of the $j$th model. The predicted value for an ensemble of $K$ models is calculated as:
\begin{equation}
\label{eqq}
{\bf{\hat y}} = {\varphi ^{ - 1}}\sum\limits_{j = 1}^K {{w_j}{{{\bf{\hat y}}}_j}} 
\end{equation}
Moreover, the variance of the ensemble  is 
\begin{equation}
\begin{array}{l}
\sigma^2 ({\varphi ^{ - 1}}\sum\limits_{i = 1}^K {{w_i}{{{\bf{\hat y}}}_i})}  = \sum\limits_{i = 1}^K {{\varphi ^{ - 2}}w_i^2} \sigma^2 ({{{\bf{\hat y}}}_i}) + 2\sum\limits_{1 \le i}\!\!\!\! {\sum\limits_{ < j \le K} {{\varphi ^{ - 2}}{\bf{\Lambda }}_i^j} } \\
= {\varphi ^{ - 1}} + 2{\varphi ^{ - 2}}\sum\limits_{1 \le i} {\sum\limits_{ < j \le K} {{\bf{\Lambda }}_i^j} } \,\,\,\,\,\forall k \in K
\end{array}
\label{eq}
\end{equation}
where $\varphi  \buildrel \Delta \over = \sum\limits_{j = 1}^K {{w_j}}$, ${\bf{\Lambda }}_i^j \buildrel \Delta \over = {w_i}{w_j}{\rm{cov}}({{{\bf{\hat y}}}_i},{{{\bf{\hat y}}}_j})$. In order to calculate the variance of the ensemble model, we should first find the variance of each specific model, i.e., $\sigma_j^2$, $j \in {\cal{P}}$ for calculating the first term in (\ref{eq}) and then calculate the covariance of the predicted values of the $i$th and $j$th model for calculating the second term in (\ref{eq}).  
In the next subsection, we deal with finding the variances for different versions of RNN.

\subsection{Bayesian RNN}
In this subsection, we present the mathematical formulations for different versions of Bayesian RNN  (i.e., Bayesian LSTM, Bayesian simple RNN, and Bayesian GRU) and the Bayesian dense layer.  Our formulations are an extended version of  
\cite{gal2016theoretically,Gal2015}.    
Consider an NN with $L$ RNN layers and $D$ dense layers.
For the ${RNN}_k$ for $\forall {\rm{ }}k \in {\cal P}$, we randomly choose one of the structures such as simple RNN, GRU, and LSTM for each hidden layer and then for each dense layer.
Let us consider  $\bf{\hat y}$ as a  predicted value of the NN model for the input $\bf{\hat x}$. Assume that the training set is ${\cal{T}} = \{ ({{\bf{X}}_{0:t}},{{\bf{y}}_t})|t \le T\}$ where ${{\bf{X}}_{0:t}}$ is the feature matrix  corresponding to the target matrix ${{\bf{y}}_t}$.  The total number of neurons at the input and output layers are $T_i$ and $T_o$, respectively. For an NN  with parameters   ${\boldsymbol \theta} $  and prior distribution $p(\bf{\boldsymbol \theta})$, we have
\begin{equation}
    \label{eq6}
p\left( {\widehat {\bf{y}}\left| {\widehat {\bf{x}},\cal{T}} \right.} \right) = \int {p\left( {\widehat {\bf{y}}\left| {\widehat {\bf{x}},{\boldsymbol{\theta }}} \right.} \right)} p\left( {{\boldsymbol{\theta }}\left| \cal{T} \right.} \right)d{\boldsymbol{\theta }}
\end{equation}
where $\hat{{\bf x}} \in \mathbb{R}^{T_i}$ is the input batch, $\hat{{\bf y}} \in \mathbb{R}^{T_o}$ is  the output and $p({\boldsymbol{\theta}}|{\bf \cal{T}})$ is the  posterior distribution. 
 As seen in (\ref{eq6}),  we have to calculate an infinite ensemble of models which is computationally impossible. That is the reason why the BNN problem cannot be solved in its original form.  
One possible solution can be finding a distribution like  $q(\boldsymbol \theta)$ that is  close  to $p\left({\boldsymbol \theta \left| {\cal{T}} \right.} \right)$. A distance metric between two distributions is Kullback-Leibler (KL) which is:
 \begin{equation}
  {\rm{KL}}(q({\boldsymbol{\theta }})\left\| {p\left( {{\boldsymbol{\theta }}\left| {\cal{T}} \right.} \right)} \right)=\sum _{{\boldsymbol{\theta }}}q({\boldsymbol{\theta }})\log ({\frac {q({\boldsymbol{\theta }})}{{p\left( {{\boldsymbol{\theta }}\left| {\cal{T}} \right.} \right)}}}).    
 \end{equation}
So the primary goal of our problem is to find a  $q({\boldsymbol{\theta }})$ for solving the following optimization problem: 
    \begin{equation}
    \label{ww}
        \mathop{{\rm{Minimize}}}\limits_{q({\boldsymbol{\theta }})} \,\,\,\,\,{\rm{   KL}}(q({\boldsymbol{\theta }})\left\| {p\left( {{\boldsymbol{\theta }}\left| {\cal{T}} \right.} \right)} \right.)   
    \end{equation}
we can rewrite (\ref{ww}) as:    
\begin{equation}
     \label{awe}
  \mathop {{\rm{Minimize}}}\limits_{q({\bf{\theta }})} \,\, - {\cal{L}}_1 + {{\cal{L}}_2}  
\end{equation}
where ${\cal{L}}_1=\!\int {q({\boldsymbol{\theta }})\log p({\bf{y}}\left| {\bf{x}} \right.,{\boldsymbol{\theta }})d{\boldsymbol{\theta }}}$, and ${{\cal{L}}_2}={\rm{KL}}(q({\boldsymbol{\theta }})\|p({\boldsymbol{\theta }}))$. As seen in equation (\ref{awe}),  the  term ${{\cal{L}}_2} $ is  the regularization
term which essentially makes the NN robust against over-fitting. The first term can be simplified by using MC sampling as $\,\,-\frac{1}{M}\,\, \sum\limits_{n = 1}^M \int{\,\log p({{\bf{y}}_n}|{h^{\boldsymbol{\theta}}({\bf{x}}_n}}))d{\boldsymbol{\theta}}$. The function ${h^{\boldsymbol{\theta}}({\bf{x}}_n})$  essentially depends on the NN structure.  It is shown in \cite{gal2016theoretically} that the second term for each  MC sample of (\ref{awe}) can be rewritten as:  
\begin{equation}
\varphi ({{{{\bf{y}}_n}},{\rm {E}}({\bf{ y}}_n| {{{\bf{x}}_n},{{{\boldsymbol{\hat \theta }}}_n}})} ) =  - {\rm{log}}{\mkern 1mu} {\mkern 1mu} p(\left. {{{\bf{y}}_n}} \right|{{\bf{x}}_n}{\rm{ }},{{{\boldsymbol{\hat \theta }}}_n})
\end{equation}
where ${\rm {E}}({\bf{ y}}_n| {{{\bf{x}}_n},{{{\boldsymbol{\hat \theta }}}_n}})$ and $\varphi $ are the output of the NN and the loss function, respectively. 
\subsection{ Approximation of Bayesian RNN Using MC Dropout}

\begin{figure}
    \centering
    \includegraphics[width=0.85\linewidth]{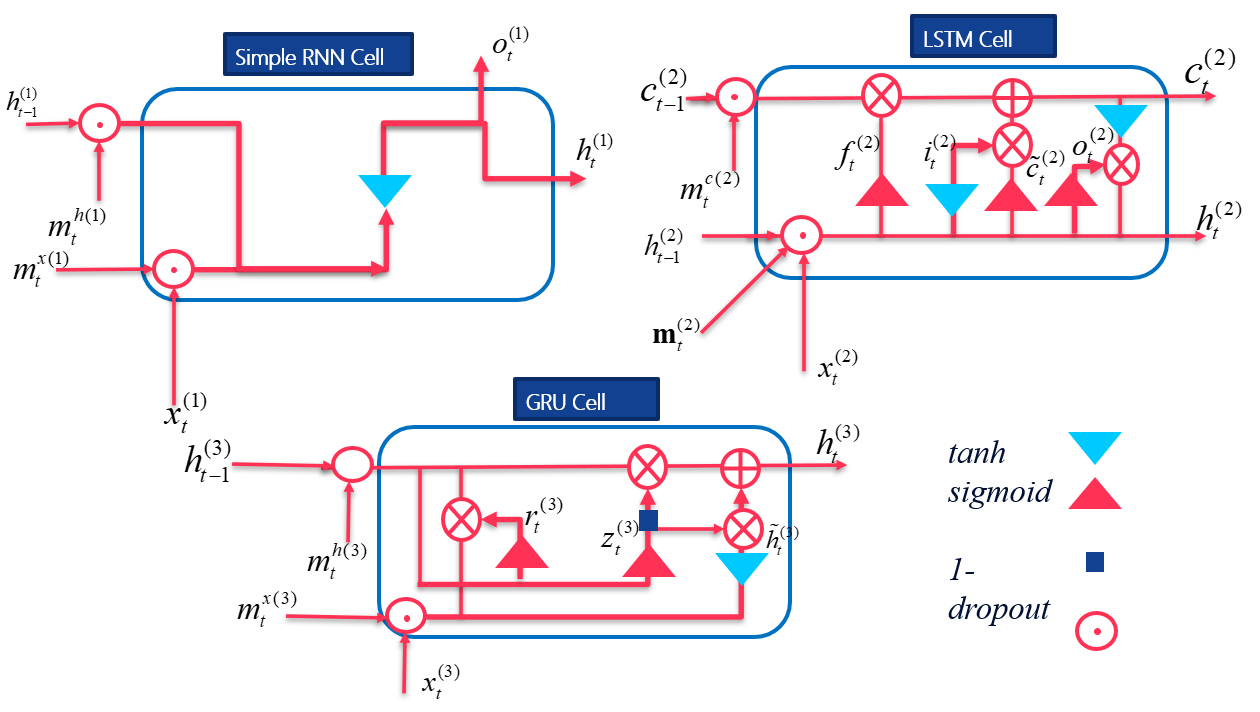}
    \caption{Different versions of RNN cells by using dropout ${\bf{m}}^{(i)}_t \forall i \in \{1,2,3\} $ at their input.}
    \label{fig3}\vspace{-20pt}
\end{figure}

Different versions of Bayesian RNN are depicted in Fig. \ref{fig3}. Let us discuss the mathematical formulations after applying the dropout mask. Considering ${\bf{h}}_t^{(i)}$ and ${\bf{x}}_t^{(i)}$  
as the hidden layer, and input of $i$th RNN scheme, let us denote ${{\boldsymbol \zeta} _t}^{(i)} = {\left[ {\begin{array}{*{20}{c}}
{{{\bf{x}}_t}}^{(i)}&{{{\bf{h}}_{t - 1}}^{(i)}}
\end{array}} \right]^T}$ and ${\bf{m}}_t^{(i)} = [m_t^{x(i)}\,\,\,m_t^{h(i)}]$.  In our notation,  we use superscripts (1), (2),  (3), and (D) for simple  Bayesian RNN, Bayesian LSTM, Bayesian GRU, and dense layers respectively. 
 $sigmoid$  and $\odot$  stand for the sigmoid function and the Hadamard  product, respectively. Moreover,  ${\bf{m}}_t^{(i)} \sim {\textit{Bernoulli}}(p^{(i)})$  is a random mask vector with a \textit{Bernoulli} distribution with probability $p^{(i)}$. 
The first topology in  Fig. \ref{fig3} is the simple Bayesian RNN, we have: 
\begin{equation}
\label{eq9}
 {\begin{aligned}{\bf h}_{t}^{(1)}&=\text{sigmoid}({{{\boldsymbol{\Sigma}}_h^{(1)}}{{\boldsymbol{\zeta}}_t^{(1)}}\odot{\bf{m}}_t^{(1)} + {{\bf{b}}_h^{(1)}}} )\\{\bf{y}}_{t}&={\boldsymbol{\Sigma}}_{y}^{(1)}{\bf h}_{t}\odot{\bf{m}}_t^{(1)}+{\bf{b}}_{y}^{(1)}\end{aligned}}  
\end{equation}
So the parameters of a  simple RNN cell are   ${\boldsymbol{\theta}}^{(1)}= [{\boldsymbol{\Sigma}}_h^{(1)},{\boldsymbol{\Sigma}}_{y}^{(1)},{\bf b}_h^{(1)},{\bf b}_y^{(1)} ]$.  
On the other hand, for approximation of Bayesian LSTM
using  dropout mask ${\bf{m}}_t^{(2)}$, we have:
\begin{align}
\label{eq12}
 {{\bf{i}}_t^{(2)}} &= \text{sigmoid} \left( {{{\boldsymbol{\Sigma}}_i^{(2)}}({{\boldsymbol \zeta} _t^{(2)}} \odot {\bf{m}}_t^{(2)})}+{{\bf b}_i^{(2)}} \right)\\
 {{\bf{f}}_t^{(2)}} &= \text{sigmoid} \left( {{{\boldsymbol{\Sigma}}_f^{(2)}}({{\boldsymbol \zeta} _t^{(2)}} \odot {\bf{m}}_t^{(2)})+{{\bf b}_f^{(2)}}} \right)\\
 {{\bf{o}}_t^{(2)}} &= \text{sigmoid} \left( {{{\boldsymbol{\Sigma}}_o^{(2)}}({{\boldsymbol \zeta^{(2)}} _t}\odot {\bf{m}}_t^{(2)})+{{\bf b}_o^{(2)}}} \right)\\
 {{{\bf{\tilde c}}}_t^{(2)}} &= \text{tanh} \left( {{{\boldsymbol{\Sigma}}_{\tilde c}^{(2)}}({{\boldsymbol \zeta} _t^{(2)}} \odot{\bf{m}}_t^{(2)})+{{\bf b}_c^{(2)}}} \right)\\
 {{\bf{c}}_t^{(2)}} &= {{\bf{f}}_t^{(2)}} \odot {{\bf{c}}_{t - 1}^{(2)}}\odot{{m}}_t^{x(2)} + {{\bf{i}}_t^{(2)}} \odot {{{\bf{\tilde c}}}_t^{(2)}}\\
  {{\bf{h}}_t^{(2)}} &= {{\bf{o}}_t^{(2)}} \odot \text{tanh}({{\bf{c}}_t^{(2)}})
\end{align}
where $ {{\bf{f}}_t^{(2)}} $,${{\bf{i}}_t^{(2)}}$, ${{\bf{c}}_t^{(2)}}$,  ${{\bf{o}}_t^{(2)}}$,  respectively are the  forget gate, input gate, LSTM unit, and output gate state values. Here, ${{\bf{h}}_t^{(2)}}$ is the state value of the LSTM gate.   
For the LSTM formulation in the first term of (\ref{awe}), we can write  ${{\bf{h}}_t^{(2)}} = {h_h^{(2)}}({{\boldsymbol{\zeta} _t^{(2)}}})$. In this case, the NN parameter is  ${\boldsymbol{\theta}}^{(2)}= [{\boldsymbol{\Sigma}}_i^{(2)}, {\boldsymbol{\Sigma}}_f^{(2)}, {\boldsymbol{\Sigma}}_o^{(2)}, {\boldsymbol{\Sigma}}_{\tilde c}^{(2)},{\bf b}_f^{(2)},{\bf b}_o^{(2)},{\bf b}_c ^{(2)}]$. 

 A GRU unit consisting of reset  and update gates is shown in Fig. \ref{fig3}. Let us briefly describe the formulations: 
\begin{equation}
{{\begin{aligned}{\bf z}_{t}^{(3)}&=\text{sigmoid}({\boldsymbol{\Sigma}}^{(3)}_{z}{\boldsymbol{\zeta} }_t^{(3)}\odot {\bf m}^{(3)}_t+{\bf{b}}_{z}^{(3)})\\ 
{\bf r}_{t}^{(3)}&=\text{sigmoid}({\boldsymbol{\Sigma}}_{r}^{(3)}{\boldsymbol{\zeta}}_t^{(3)}\odot {\bf m}^{(3)}_t+{\bf{b}}_{r}^{(3)})
\\
{\cal{Y}}_t^{(3)}  & \buildrel \Delta \over =  {{\boldsymbol{\Sigma}}_{h}^{(3)}}{\boldsymbol{\zeta }_t}^{(3)}\odot {\bf m}^{(3)}_t+{\bf U}_{h}^{(3)}({\bf{r}}_{t}^{(3)}\odot {\bf{h}}_{t-1}^{(3)}\odot {\bf m}^{{\bf{h}}(3)}_t)+{\bf{b}}^{(3)}_{h}\\{\bf\hat  {h}}_{t}^{(3)}&=\text{tanh}({\cal{Y}}_t^{(3)})
\\{\bf h}_{t}^{(3)}&={\bf z}_{t}^{(3)}\odot {\bf \hat {h}}^{(3)}_{t}\odot {\bf m}^{(3)}_t+(1-{\bf z}_{t}^{(3)})\odot {\bf h}_{t-1}^{(3)}\odot {\bf m}^{(3)}_t\end{aligned}}}
\end{equation}

 In this case, the NN parameter is  ${\boldsymbol{\theta}}^{(3)}= [{\boldsymbol{\Sigma}}_h^{(3)}, {\boldsymbol{\Sigma}}_r^{(3)}, {\boldsymbol{\Sigma}}_z^{(3)}, {\boldsymbol{U}}_{h}^{(3)},{\bf b}_r^{(3)},{\bf b}_z^{(3)},{\bf b}_h ^{(3)}]$. 
 Moreover,  if we apply a linear layer after the recurrent cells, we should apply a dropout mask at the linear layers as well. Given  ${\bf b}_k$ as the bias of the $k$th linear layer,  $k \in \cal{K}$, ${\cal{K}}=\{1,...,L\}$, where $L$ is the total number of linear (dense) layers.   We drop the neuron randomly by \textit{Bernoulli} distribution, i.e.,  
\begin{align}
\label{eq17}
{{\boldsymbol{\Sigma}}_k^{(D)}} &= {\rm{Diagonal}}({[{b_{k,j}}]_{j = 1}^{{K_i}}}){{\bf{T}}_k^{D}}\\ \nonumber
{b_{k,j}^{(D)}} &\sim {\textit{Bernoulli}}\left({{p_k^{(D)}}} \right)\,\,\,\,\,\forall k \in {\cal{K}}\,,\,j  \in \cal{J}
\end{align}
where ${\cal{J}}=\{1,...,{n^{k-1}}\}$ is the dimension of $(k-1)$th unit, $p_k^{(D)}$ is the probability of dropping out, and matrix ${{\bf T} _k^{(D)}}$ is the NN variational
parameters. If we drop out the $j$th unit of $(k-1)$th layer which is digested to $k$th layer, the binary variable ${b_{k,j}}= 0$.
Assume that ${{\boldsymbol{\hat \theta }}_n} \sim {\rm{ }}q\left({\boldsymbol{\theta }} \right)$ is an $n$th random response of the NN following the MC dropout employment. The mean and variance then can be calculated as \cite{Gal2015}:
\begin{equation}
\label{eq18}
{{\rm E}_{q({\bf{y}}  \left| {{\bf{ x}} } \right.)}}({{\bf{y}}}) \approx \frac{1}{M}\sum\limits_{m = 1}^M {{{\rm E({\bf{  y}}}}|{{\bf{ x}}},{\bf{{ \boldsymbol \theta}}}^m})
\end{equation}
\begin{equation}
\label{eq19}
{{\rm E}_{q({\bf{y}}  \left| {{\bf{ x}} } \right.)}}({ {{{\bf{ y}}}}^T} {{{\bf{ y}}}}) \approx 
  \frac{1}{M}\sum\limits_{m = 1}^M {{{{\rm E(\bf{y}}|}}{{{{\bf{ x}}},{\bf{ {\boldsymbol \theta}}}^m)}^T}} {{\rm E({\bf{ y}}|}}{{\bf{ x}}},{\bf{ {\boldsymbol \theta}}}^m)
\end{equation}
The model’s predictive variance is
\begin{equation}
\label{eq20}
 {\sigma_{q({\bf{y}}  \left| {{\bf{ x}} } \right.)} }({\bf{y}}) \!=\! {{\rm{E}}_{q({\bf{y}}  \left| {{\bf{ x}} } \right.)} }({{\bf{y}}^T}{\bf{y}}) \!-\! {{\rm{E}}_{q({\bf{y}}  \left| {{\bf{ x}} } \right.)} }{({\bf{y}})^T}{{\rm{E}}_{q({\bf{y}}  \left| {{\bf{ x}} } \right.)} }({\bf{y}})  
\end{equation}
The following algorithm briefly shows the MC dropout steps. 

\renewcommand{\baselinestretch}{0.85}{
\begin{algorithm}[H]
\hspace*{\algorithmicindent} 
\begin{algorithmic}[1]\small
\caption{Bayesian approximation using MC Dropout}
\State{\textbf{Input}:  ${\theta^{(i)}\forall i\in\{1,2,3,D\}}$ }
\State{\textbf{Output}:  model mean and variance ${{\rm E}_{q({\bf{y}}  \left| {{\bf{ x}} } \right.)}},  {\sigma_{q({\bf{y}}  \left| {{\bf{ x}} } \right.)} } $}
\While { $m \le N_{dropout}$ }:
\State{   Generate \textit{Bernoulli} random samples and set equations (\ref{eq9})-(\ref{eq17}) to find  
${\boldsymbol{\hat \theta^{(i)}}}$ 
where ${\boldsymbol{\hat \theta^{(i)}}}\sim q({\boldsymbol{ \theta^{(i)}}})$. and mean and variance in (\ref{eq18})-(\ref{eq20})}
\State{$m=m+1$}
\EndWhile
    \State {Solve the minimization problem in  ({\ref{awe}}).}
\end{algorithmic}
\end{algorithm}}

\renewcommand{\baselinestretch}{0.85}{
\begin{algorithm}\small
\hspace*{\algorithmicindent} Assume that the number of iterations is a random number between ${NI}_{min}$ and ${NI}_{max}$ and the total number of populations is 
a random number between ${NP}_{min}$ and ${NP}_{max}$. 
\hspace*{\algorithmicindent} 
\begin{algorithmic}[1]
\caption{VGE steps}
\State{\textbf{Input}: ${\bf x}(t)$, and $j$th model mean and variance i.e. ${{\rm E}_{q({\bf{y}}  \left| {{\bf{ x}} } \right.)}^{(j)}},  {\sigma_{q({\bf{y}}  \left| {{\bf{ x}} } \right.)}^{(j)} } $},  $\forall j \in \{1,...,K\}$.
\State{\textbf{Output}: Ensemble model }

\State{$NI  \sim  U({NI}_{min},{NI}_{max} )$}
\While { $i \le NI$ }
\State{$NP \sim  U({NP}_{min},{NP}_{max} )$}
\While{$g< NP$}
\For{$j\le 1 \le K$}
\State{Generate initial population $ {{{RNN}_j}}$}
\State{Number of layers: $L \sim  U({L}_{min},{L}_{max} )$}
\State{Number of units: $N \sim  U({N}_{min},{N}_{max} )$}
\State{Generate the $ {RNN_j} $ }

\State{Fitness and evaluate $ {{{RNN}_j}} $}
\If{$F_1^{(g)}$ > $F_1^{(opt)}$ }
\State{$RNN_j^{(opt)}$= $RNN_j^{(g)}$}
\EndIf
\State{Create offspring using crossover}
\State{Mutate offspring}
\State{ $g=g+1$}
\EndFor
\EndWhile
\State{  $i=i+1$}
\EndWhile
\end{algorithmic}
\end{algorithm}}
\section{Results}
In this section, we have evaluated the proposed VGE AD approach. All the tests were conducted on the NVIDIA GeForce GTX
1080Ti GPU. We used Keras with the TensorFlow backend for implementations. First, we describe our model structure, hyperparameters, settings, and methodology; then, we compare its performance with
various benchmarks. The source code can be found at \url{http://bit.ly/3YDCgDJ}.

\subsection{Methodology}

Consider $K=2$ RNNs with the minimum and the maximum number of layers $L_{min}=2$ and $L_{max}=6$, respectively. 
The minimum and the maximum number of units are  $N_{min}=128$ and $N_{max}=256$, respectively. The batch size is supposed to be $256$, and we use `ADAM' as an optimizer. The maximum dropout rate is 0.2 and we run the simulations for  the total number of 100 epochs.   The total  number of samples for MC dropout is  1000. 
Moreover, the mutation rate is 0.1, the minimum mutation is 0.0001, the momentum is 0.1,  and the maximum mutation momentum is 0.1. The minimum and the maximum  populations are $NP_{min}=4$ and  $NP_{max}=6$ respectively. 
The minimum and maximum number of iterations are $NI_{min}=3$ and $NI_{max}=6$, respectively. 
 Moreover,  the following  approaches are compared:  
     (i)  Bayesian LSTM approach in   \cite{2022};
  (ii)   MadGan approach in \cite{li2019mad}; 
  (iii)  LSTM with dynamic thresholding approach in    \cite{hundman2018detecting};
  (iv)  TadGan method in \cite{geiger2020tadgan}; 
  (v)  Arima approach in  \cite{pena2013anomaly}; 
 (vi) LSTM approach in  \cite{malhotra2015long};
 and (vii) Our VGE approach.
  
To lessen the false positives (FP), we use a  post-processing procedure to determine anomalies.  In our dataset, anomalies often take place in a sequential sample. In this case, when we evaluate the fitness population of the GA for each of the RNNs, if an anomaly is detected, we wait until a certain number of tentative points occur and then flag the whole sequence as detected anomalies. Setting the maximum waiting time for flagging the anomaly sequence ($\tau_{max}$) is an important step. 
Indeed, this  post-processing significantly improves the performance. Although by increasing the value of $\tau_{max}$ up to a specific threshold,  we can get better performance, it increases the overall judgment time for AD. 
Indeed, the best value for $\tau_{max}$ can be achieved by compromising online implementation and accuracy. In the next subsection, we  discuss the importance of selecting  the optimal value for $\tau_{max}$, i.e. ($\tau_{max}^{opt}$), in detail.  Moreover, we discover that some of the signals are not useful for evaluating AD methods. For example, the telemetry value of  signal ``M6'' in the MSL dataset is -1 for the training phase. For this reason, we exclude signals ``P10, M6, E3, A1, D16,  D1, D3, F3, A6, P14,  D5,  G6, P15,  G1, M3, R1, D4,  P11,   M2,  D11'' from the evaluations.  
\subsection{Evaluation}
In this subsection, the performances of different methods are compared with VGE. Assume $TN$, $FN$, $FP$, and $TP$ as true negative, false negative, false positive, and true positive in the confusion matrix. The following  evaluation measures have been utilized to compare different methods: 

1) Mean Squared Error (MSE):
 We use MSE to evaluate how much the method is accurate for prediction. This metric returns the squared gap of the predicted  and the real values, i.e., ${\rm{MSE}} = {\left\| {{\bf{y}} - {\bf{\hat y}}} \right\|^2}$ where ${\bf{y}} = {[{y_1},...,{y_T}]^T}$ and ${\bf{\hat y}} = {[{{\hat y}_1},...,{{\hat y}_T}]^T}$ are the real and predicted vector respectively.  The MSE comparison for different models has been depicted in Table \ref{T2}. In Table \ref{T2}, 
 we have not compared methods such as MadGan and TadGan  which are reconstruction-based methods
 as they are reconstructing the TS and providing the reconstruction error which  is different from the MSE terminology here.
2) Recall:  $\eta  = \frac{{TP}}{{TP + FN}}$;  
3) Precision:  $\tau   = \frac{{TP}}{{TP + FP}}$;
4) $F_{1}$:  
${F_1^{-1}} =  {\frac{{{\tau ^{ - 1}} + {\eta ^{ - 1}}}}{2}} $.
5) Accuracy:  ${{\kappa }} = \frac{{TP + TN}}{{TP + TN + FP + FN}}$.
 
\begin{figure}
    \centering\includegraphics[width=\linewidth]{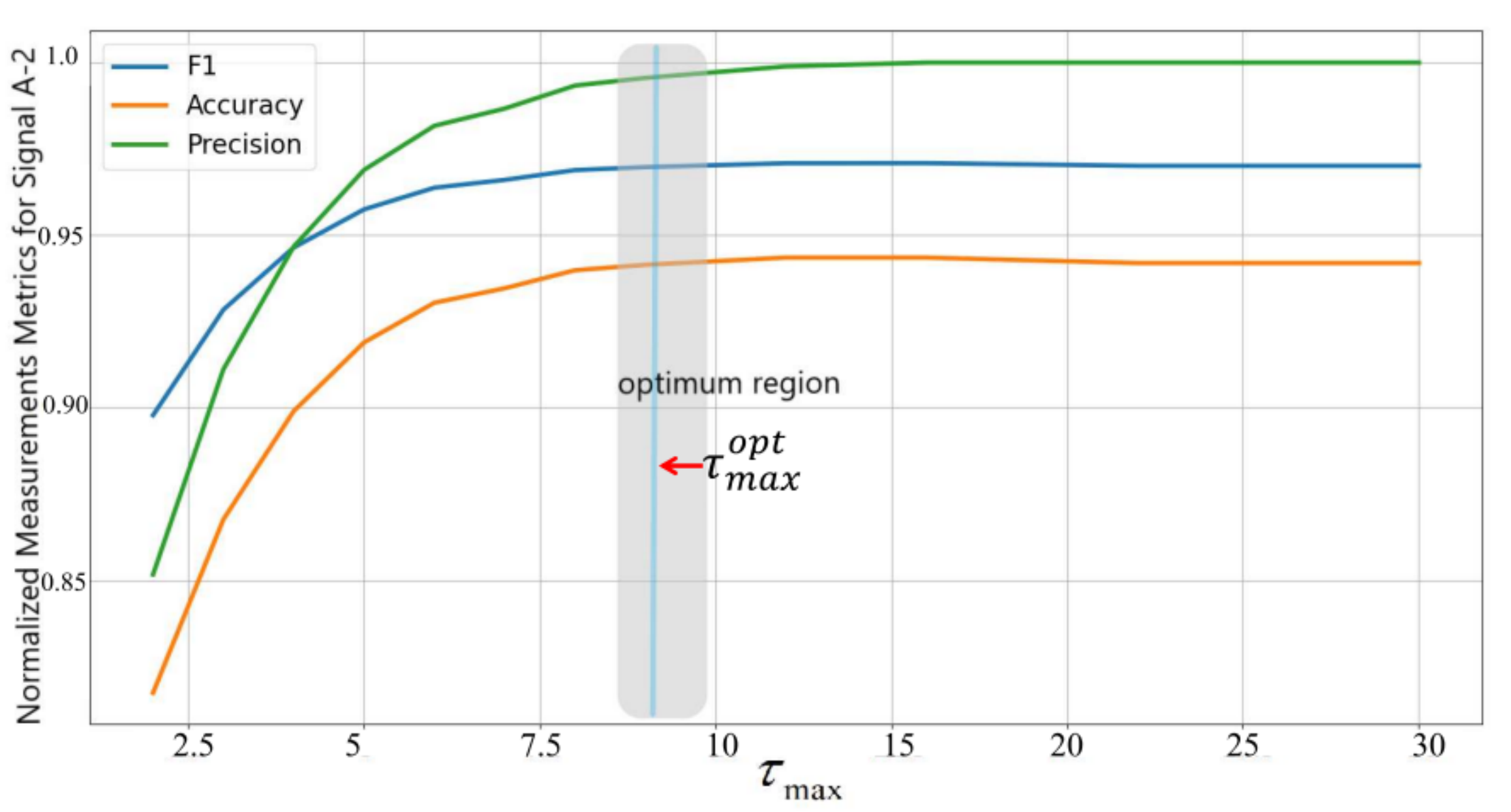}
    \caption{ $\tau_{max}$ effect in evaluation for signal `A-2' in VGE}
    \label{Fig6}
\end{figure}

Let us discuss how to choose the optimal value of
$\tau_{max}$ in post-processing. We discover that the value of $\tau_{max}$ can make a huge difference in the final results. For this reason, our approach is to apply a grid search for $\tau_{max}$ and pick the optimal value. 
 As shown in Fig. \ref{Fig6}, the normalized values of $F_1$, Accuracy, and Precision for VGE  have been depicted. We draw this figure for just one of the signals, i.e., `A-2', but it can be extended to other signals.  As seen in this figure, after a particular value of $\tau_{max}$, the results do not change, and as a result, that point is considered as an optimal point. For 'A-2' this specific optimal  point is $\tau^{opt}_{max}=9$. 
 \begin{table}[!ht]
\centering
   \caption{Mean Square Estimation comparison }
 \scalebox{0.93}{\begin{tabular}{|c|c|c|c|c|}
    \hline
   Baseline & Bayesian LSTM \cite{2022} &   LSTM \cite{malhotra2015long}& Arima \cite{pena2013anomaly}&\textbf{VGE} \\ \hline
      SMAP  &{0.05}& {0.07}&0.61&\textbf{0.02}\\ \hline
      MSL &{0.21}&  0.29&0.83&\textbf{0.09}\\
      \hline
    \end{tabular}
   }
    \label{T2}
\end{table}

As seen in Table \ref{T222}, the ${F_1}$ score of different approaches are compared. As the results of the comparison show, using VGA can significantly improve the ${F_1}$ score.  Given that the  MadGan method has significantly lower performance than the other approaches, we have eliminated this method from the rest of our comparisons. 

Despite improving the  $F_1$ score, our VGE method poses an extra complexity to the overall training process. The performance and complexity trade-off should be considered when designing the system.  The complexity of different methods in terms of training and testing times  has been derived in Table \ref{T4}.  For average values of ${ \overline{NI}}=4$ and ${\overline {NP}}=5$, the training time of the VGE method would be $96.4$ seconds which is higher than other approaches. 
Note that, the use of GA is the computational bottleneck of our algorithm. By using another low complexity evolutionary algorithm such as \cite{deng2022enhanced}, the training time can be reduced. 
 By assigning smaller values to  ${ \overline{NI}}$, and ${\overline {NP}}$, the training and prediction times  will be reduced, in a cost of slightly worse performance. Depending on the use cases, and the trade-off between time complexity and performance,   the optimal values for ${ \overline{NI}}$, and ${\overline {NP}}$ are selected.  
 In Table \ref{tableV}, accuracy, precision, and recall for two datasets, and different methods are derived. As Table \ref{tableV} shows, for both datasets, our VGE method outperforms other methods.
\begin{table}
\centering
   \caption{Evaluation of different methods by F1 score }    
 \scalebox{0.55}{\begin{tabular}{|c|c|c|c|c|c|c|c|c|c|}
    \hline
   \large{Baseline} & \large{Bayesian LSTM}  & \large {TadGan}  &\large{Arima}  &\large{LSTM AE} & \large{MadGan}&\large{LSTM }& \large{\textbf{ VGE}}\\ \hline
      \large{SMAP}  &\large{0.84}& \large{0.66}&\large{0.42}&\large{0.69}&\large{0.12}&\large{0.62}&\large{\textbf{0.92}}\\ \hline
      \large{MSL} &\large{0.74}&  \large{0.55}&\large{0.49}&\large{0.55}&\large{0.11}&\large{0.48}&\large{\textbf{0.84}}\\
      \hline
    \end{tabular}
    }
    \label{T222} \vspace{-50pt}
\end{table}

\begin{table}[ht!]\vspace{-10pt}
\centering
\caption{Comparison of time complexity of different methods}
   \scalebox{0.9}{ \begin{tabular}{|c|c|c|c|}
    \hline
    Methods &  { Training Time} (sec per epoch) &  Throughput\\ \hline
    Bayesian LSTM\cite{2022} &  {1.48} &  $0.57 \times 10^3$ \\
      \hline
      TadGan \cite{geiger2020tadgan} & {$78$} & {$0.27\times 10^2$} \\ \hline
      Arima \cite{pena2013anomaly} & 1.1  & {$2.1\times 10^3$}  \\
      \hline
     LSTM AE \cite{hundman2018detecting} &  1.84   & $1.16 \times 10^3$ \\ \hline
          LSTM \cite{malhotra2015long} & {1.48}  & {$1.1\times 10^3$}  \\ \hline
                  \textbf{VGE}  &   \textbf{96.4}  &  \textbf{0.54 $\times$  {$\bf 10^2$}}  \\ \hline
    \end{tabular}}
   \label{T4}
\end{table}

\begin{table}[ht!]\vspace{-20pt}
\centering
\caption{Comparison of accuracy, precision, and recall}
\label{tableV}
\scalebox{0.79}{
\centering
\begin{tabular}{|c|c|c|c||c|c|c|}
\hline
\multicolumn{1}{|c|}{\bfseries  Methods} &
\multicolumn{3}{c||}{\bfseries SMAP}&
\multicolumn{3}{c|}{\bfseries  MSL}\\
{} &  Accuracy &   Precision &  Recall &  Accuracy &   Precision &  Recall\\
\hline
{Bayesian LSTM \cite{2022}} & {0.75} & {0.82} &{ 0.87} & {0.7} & {0.74} & {{0.87}}\\
\hline
{TadGan}{\cite{geiger2020tadgan} } & {0.76} & {0.76} & {0.69} & {0.58} & {0.58} & {0.68}\\
\hline
{Arima\cite{pena2013anomaly}} & {0.52} & {0.61} & {0.57} & {0.49} & {0.63} & {0.45} \\
\hline
{LSTM AE\cite{hundman2018detecting}} &  0.73 & 0.76 & {0.75} &  0.55 &  0.65 & 0.66\\
\hline
{LSTM \cite{malhotra2015long}} & 0.67  &  0.56 &  0.73 & 0.53  & {0.52} &{0.67}\\
\hline
 \textbf{VGE } & \textbf {0.92}  &  \textbf{0.89}&  \textbf{0.92} & \textbf{0.79}  &  \textbf{0.8} & \textbf{0.89}\\
\hline
\end{tabular}
}
\end{table}

\section{Conclusion}
In this paper,  we propose a variance-based ensemble of  various Bayesian NNs for satellite telemetry channel AD which was  tuned by GA for  optimal topology selection. 
Finding the variance is challenging, and we utilize the MC dropout approach for finding each model's variance. After prediction, we use a post-processing method for more efficient AD. Simulation results confirm a higher performance in terms of prediction and AD than traditional methods. Nevertheless, our approach introduces time complexity, and the real-time implementations of our algorithm remain an interesting avenue for future work.
Moreover, our method can be applied to the spacecraft's FDIR and additional multivariate TS AD tasks. 

\bibliographystyle{IEEEtran}
\bibliography{ref.bib}

\end{document}